\relax
\documentclass[letterpaper]{article} 
\usepackage{aaai21}  
\usepackage{times}  
\usepackage{helvet} 
\usepackage{courier}  
\usepackage[hyphens]{url}  
\usepackage{graphicx} 
\usepackage{algorithm}
\usepackage{algorithmic}
\usepackage{amssymb}
\usepackage{float}
\usepackage{amsmath}
\usepackage[backref]{hyperref} 

\usepackage{epsfig}
\usepackage{graphicx}
\usepackage{multirow}
\usepackage{tabularx}
\usepackage{booktabs}
\usepackage{url}
\usepackage{makecell}

\usepackage{float}
\usepackage{subfigure}
\usepackage{graphicx}

\usepackage{natbib}  
\usepackage{caption} 
\title{PP-ShiTu: A Practical Lightweight Image Recognition System}
\author {
    Shengyu Wei, Ruoyu Guo, Cheng Cui, Bin Lu, Shuilong Dong, Tingquan Gao, Yuning Du, \\
    Ying Zhou, Xueying Lyu, Qiwen Liu, Xiaoguang Hu, Dianhai Yu, Yanjun Ma \\
}
\affiliations {
    Baidu Inc. \\
    \{weishengyu, zhouying15\}@baidu.com \\
}

\begin{document}

\maketitle

\begin{abstract}
In recent years, image recognition applications have developed rapidly. A large number of studies and techniques have emerged in different fields, such as face recognition, pedestrian and vehicle re-identification, landmark retrieval, and product recognition. In this paper, we propose a practical lightweight image recognition system, named PP-ShiTu, consisting of the following 3 modules,mainbody detection, feature extraction and vector search. We introduce popular strategies including metric learning, deep hash, knowledge distillation and model quantization to improve accuracy and inference speed. With strategies above, PP-ShiTu works well in different scenarios with a set of models trained on a mixed dataset. Experiments on different datasets and benchmarks show that the system is widely effective in different domains of image recognition. All the above mentioned models are open-sourced and the code is available in the GitHub repository PaddleClas\footnote{https://github.com/PaddlePaddle/PaddleClas} on PaddlePaddle\footnote{https://github.com/PaddlePaddle/Paddle}.
\end{abstract}

\section{Introduction}
Image recognition is a general task of computer vision. Benefiting from the rapid improvement of deep learning and the huge market demand, image recognition has grown rapidly, and developed into several sub-fields. Many applications use similar pipelines, such as face recognition, pedestrian and vehicle re-identification, landmark recognition, product recognition, etc. In addition, strategies that work in one application often prove to be effective for other applications as well. For example, ArcMargin loss \cite{deng2019arcface} and Triplet loss \cite{wen2016discriminative} were proposed for face recognition, which was shown to be helpful for many other tasks such as pedestrian re-identification \cite{hermans2017defense} and landmark recognition \cite{yokoo2020two}. 
\begin{figure}[t]
\centering
\subfigure{
\begin{minipage}[b]{1\linewidth}
\centering
\includegraphics[width=\columnwidth]{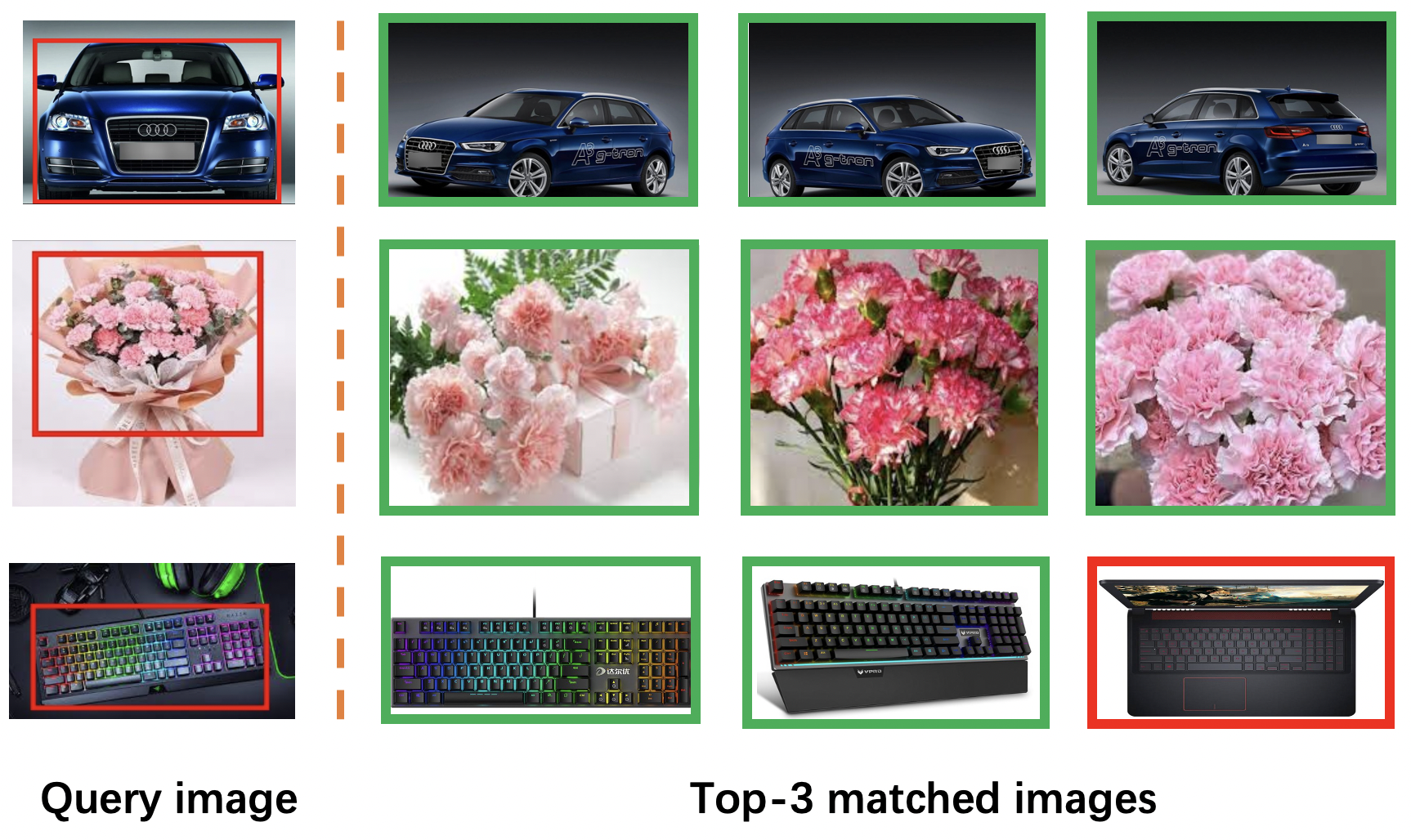}
\end{minipage}
}

\caption{The top-3 results returned by the the proposed PP-ShiTu system for different queries. The third returned result in the third line is wrong because there are only two keyboards in the gallery dataset.}
\label{rec_demo}

\end{figure}

\begin{figure*}[t]
\centering
\includegraphics[width=15cm]{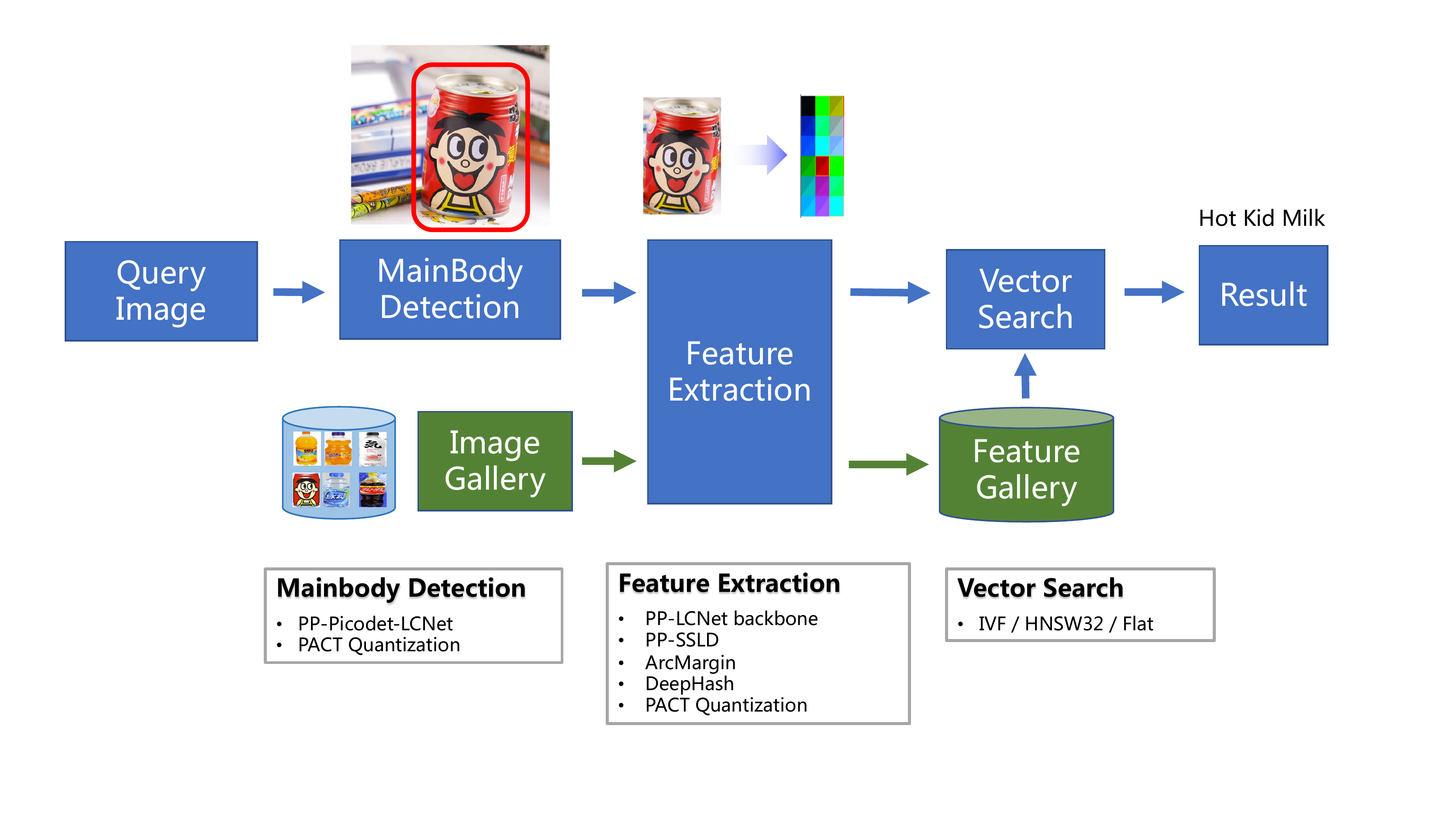}
\caption{The framework of the proposed PP-ShiTu.}
\label{framework}

\end{figure*}

However, researchers always focus on one or a few applications, hence, as open source projects. When migrating a strategy to another application,it costs a lot. In view of this, we propose the PP-ShiTu image recognition system to solve similar problems in one pipeline. Figure \ref{framework} illustrates the framework of PP-ShiTu. PP-ShiTu contains three modules, namely mainbody detection, feature extraction and vector search. The pipeline of PP-ShiTu is very simple. When getting an image, we first detect one or several mainbody areas to find the main regions of the image. Then we use a CNN model to extract the features from these regions. A feature is a floating point vector or a binary vector. According to metric learning theory, features imply the similarity of two objects. The shorter the distance between two features, the more similar the original two objects are. Finally we use a vector search algorithm to find the feature in the gallery that are closest to the features we extracted from the image and use the corresponding labels as our recognition results.

In addition to building a generic pipeline, we also introduce a number of effective strategies in the system. We use PP-Picodet with backbone of PP-LCNet \cite{cui2021pplcnet} as the backbone of the mainbody detection model. And PP-LCNet with metric learning strategy, such as ArcMargin \cite{deng2019arcface} is used as the feature extraction model. Knowledge distillation strategies are widely used in the system. We use backbones trained with SSLD distillation strategy \cite{cui2021selfsupervision} and we also use SSLD to train feature extraction models. Model quantization \cite{choi2018pact} helps us to reduce the storage size of the model. And DeepHash \cite{wu2019deep} strategy was used to compress the features so as to accelerate the vector search. The ablation experiments show the effectiveness of the strategies above.

What's more, we train the mainbody detection model and the feature extraction model with a hybrid dataset by mixing several datasets together. PP-ShiTu works well in different scenarios and obtains competitive results on many datasets and benchmarks. 

The rest of paper is organised as follows. In section 2, we present the details of the modules and strategies of PP-ShiTu. Then in section 3, we present and discuss the results of the ablation experiments. Finally, we summarize the conclusions in section 4.

\section{Modules and Strategies}

\subsection{Mainbody Detection}

The mainbody detection technology is a widely used detection technology, which refers to detecting all the foreground objects in an image. Mainbody detection is the first step of the whole recognition task, which can effectively improve the recognition accuracy.

There are a wide variety of object detection methods, such as the commonly used two-stage detectors (FasterRCNN series, etc.), single-stage detectors (YOLO, SSD, etc.), anchor-free detectors (PP-PicoDet, FCOS, etc.) and so on. PaddleDetection developed the PP-YOLOv2 models for server-side scenarios and the PP-PicoDet models for end-side scenarios (CPU and mobile), both of which are SOTA in their respective scenarios. By combining many optimization tricks such as Drop Block \cite{ghiasi2018dropblock}, Matrix NMS \cite{wang2020solov2}, IoU Loss \cite{rezatofighi2019generalized} and so on, PP-YOLOv2 exceeds YOLOv5 on both accuracy and efficiency. PP-PicoDet takes PP-LCNet as its backbone and combines many other detector training tricks such as PAN FPN \cite{liu2018path}, CSP-Net \cite{wang2020cspnet}, SimOTA \cite{ge2021yolox}, eventually becoming the first object detector with mAP(0.5:0.95) surpassing 0.30+ on the COCO dataset within 1M parameters when the input size is 416.

A lightweight mainbody detection model is required for a lightweight image recognition system.Therefore, we utilize PP-PicoDet in PP-ShiTu.

\subsubsection{Lightweight CPU Network (PP-LCNet)}

In order to get a better accuracy-speed trade-off on Intel CPU, we obtained a BaseNet based on comparing different network inferences and added different methods to further improve the accuracy of this network, resulting in the PP-LCNet, which provides a faster and more accurate recognition algorithm with mkldnn enabled. The structure of the entire network is shown in Figure \ref{PP-LCNet}. A brief description of the improvements of the network is given below. For more details and ablation experiments, please refer to the original paper \cite{cui2021pplcnet}.

\begin{figure}[h!]
\centering
\subfigure{
\centering
\includegraphics[width=\columnwidth]{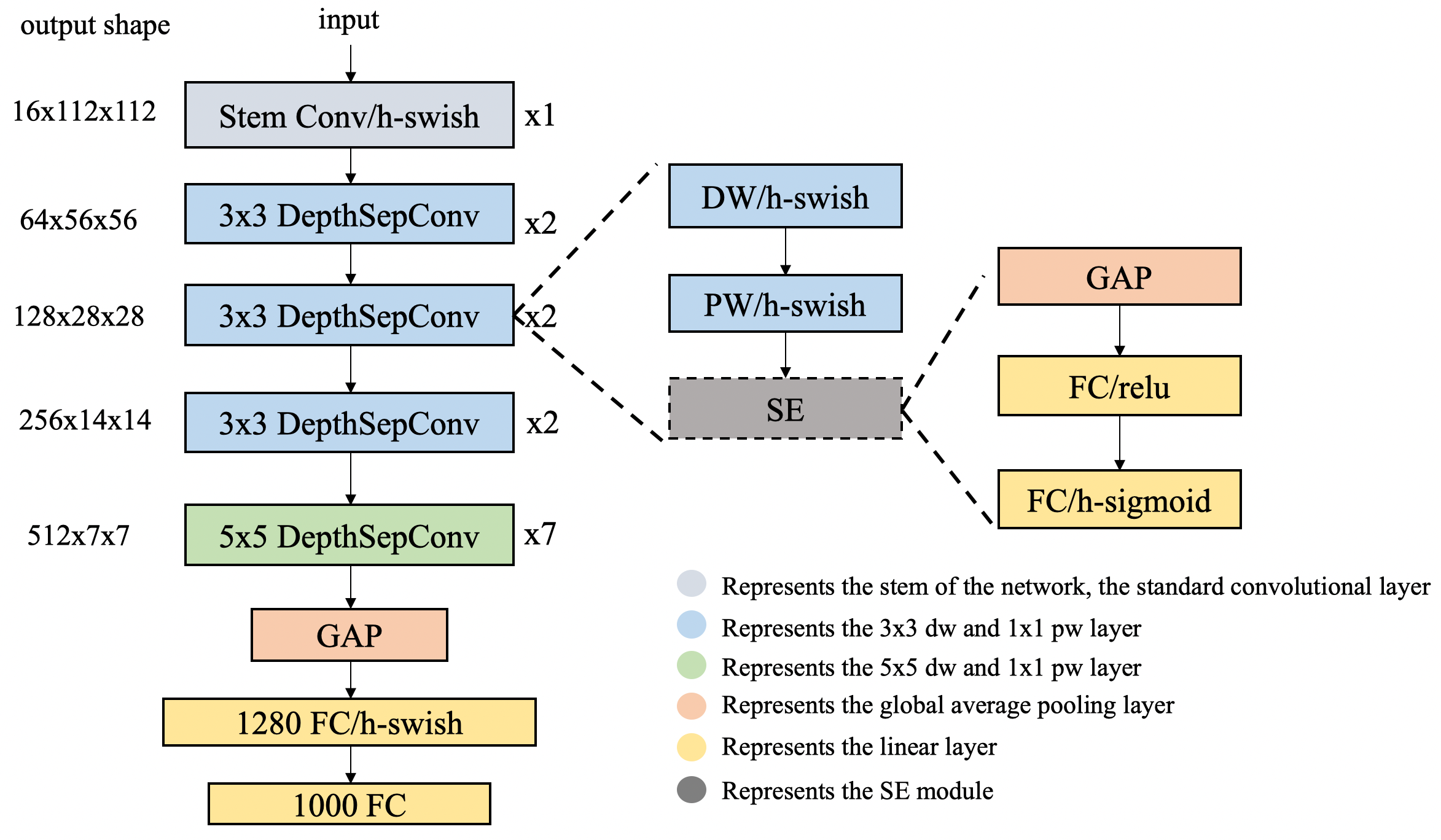}
}
\caption{PP-LCNet network structure. The dotted box represents optional modules. The stem part uses standard $3\times3$ convolution. DepthSepConv means depthwise separable convolutions, DW means depthwise convolutions, PW means pointwise convolutions, GAP means global average pooling.}
\label{PP-LCNet}
\end{figure}
 
\textbf{Better activation function.}
In order to improve the fitting ability of BaseNet, we replaced the activation function in the network with H-Swish from the original ReLU, which can bring a significant improvement in accuracy with only a slight increase in inference time.

\textbf{SE modules at appropriate positions}
SE \cite{Hu_2018_CVPR} module has been used by a large number of networks since it was proposed. It is a good way to weight the network channels to obtain better features, and is used in many lightweight networks such as MobileNetV3 \cite{Howard_2019_ICCV}. However, on Intel CPUs, the SE module increases the inference time, so we cannot use it for the whole network. In fact, through extensive experiments, we have found that the closer to the tail of the network, the more effective the SE module is. So we just add the SE module to the blocks near the tail of the network, which results in a better accuracy-speed balance. The activation functions of the two layers in the SE module are ReLU and H-Sigmoid respectively.

\textbf{Larger convolution kernels}
The size of the convolution kernel tends to affect the final performance of the network. In mixnet \cite{1907.09595}, the authors analyzed the effect of different sizes of convolution kernels on the performance of the network, and ended up mixing different sizes of convolution kernels in the same layer of the network. However, such a mixture slows down the inference speed of the model, so we tried to increase the size of convolution kernels, but minimize the increase in inference time. In the end, we set the size of the convolution kernel at the tail of the network as $5 \times 5$.

\textbf{Larger dimensional 1 × 1 conv layer after GAP}
In PP-LCNet, the output dimension of the network after GAP is small, and directly connecting the final classification layer will lose the combination of features. In order to give the network a stronger fitting ability, we insert a 1280-dimensional size 1x1 conv between GAP layer and final classification layer, which allows more storage space for the model with little increase of inference time.

With these four changes, our model performs well on ImageNet, and table \ref{PP-LCNet-Ablation} lists the metrics compared to other lightweight models on Intel CPUs.

\subsubsection{PP-PicoDet}

PP-PicoDet is a new series of object detection models developed by PaddleDetection\footnote{https://github.com/PaddlePaddle/PaddleDetection}\label{ppdet}.

Specifically, for the backbone model, we utilize the PP-LCNet described above, which helps to improve the detector's inference speed as well as mAP. For the neck, we combine CSP-Net \cite{wang2020cspnet} with PAN FPN \cite{liu2018path} to develop a new FPN structure called CSP-PAN, which helps to enhance the feature map extraction ability. SimOTA \cite{ge2021yolox} is adopted and an improved G-Focal loss \cite{li2020generalized} function is used in the network training process. What's more, some empirical improvement is also used for PP-PicoDet. For example, we use a  depthwise convolution with 5x5 kernel rather than 3x3 in the FPN since it can improve mAP by 0.5\%  and almost do no harm to the inference speed. More details can be seen on PaddleDetection.

For the mainbody detection task, all objects are seen as foreground, so there is only one class named \textbf{foreground} in the label list.

\subsection{Feature Extraction}

The main problem of image recognition is how to extract better features from the model. Therefore, the ability of feature extraction directly affects the performance of image recognition. In the training phase of feature extraction, we use the metric learning method to learn the features of an image. The metric learning is briefly described below.

\subsubsection{Additive Angular Margin Loss (ArcMargin loss)}

Metric learning attempts to map data into an embedding space, where similar data are close together and dissimilar data are far apart.
In metric learning, the quality of features depends on loss, backbone, data quality and quantity, and training strategy. Loss is the most important part of metric learning.
The loss of metric learning is divided into two types, namely loss classification-based and pair-based.
In recent years, this type of loss has been increasingly used due to improved version of the classification-based loss are more robust. In PP-ShiTu, we use ArcMargin \cite{deng2019arcface} loss, which is an improvement based on Softmax loss.
The loss of Arcmargin is shown in the formula, and the loss maximizes the classification limit in the angular space, so that the features are better extracted and organized.

\begin{equation}
{L}=-\frac{1}{N}\sum_{i=1}^{N}\log\frac{e^{s(\cos(\theta_{y_i}+m))}}{e^{s(\cos(\theta_{y_i}+m))}+\sum_{j=1,j\neq  y_i}^{n}e^{s\cos\theta_{j}}}.
\label{eq:arcface}
\vspace{-1mm}
\end{equation}

where $\theta_j$ is the angle between the weight and the feature. The batch size and the class number are $N$ and $n$, respectively. $m$ is an angular margin penalty on the target angle $\theta_{y_i}$, and $s$ is feature scale. 

\subsubsection{Unified-Deep Mutual Learning (U-DML)}

Deep mutual learning \cite{dml2017} is a method in which two student networks learn from each other, and a larger teacher network with pre-trained weights is not required for knowledge distillation. In DML, for image classification task, the loss functions contains two parts: (1) loss function between student networks and groundtruth. (2) Kullback–Leibler divergence (KL-Div) loss among the student networks' output soft labels.

In the work of OverHaul \cite{Overhaul}, in which feature map distance between student network and teacher network is used for the distillation process. Transform is carried out on student network feature map to keep the feature map aligned.

\begin{figure*}[h!]
\centering
\subfigure{
\centering
\includegraphics[width=14cm]{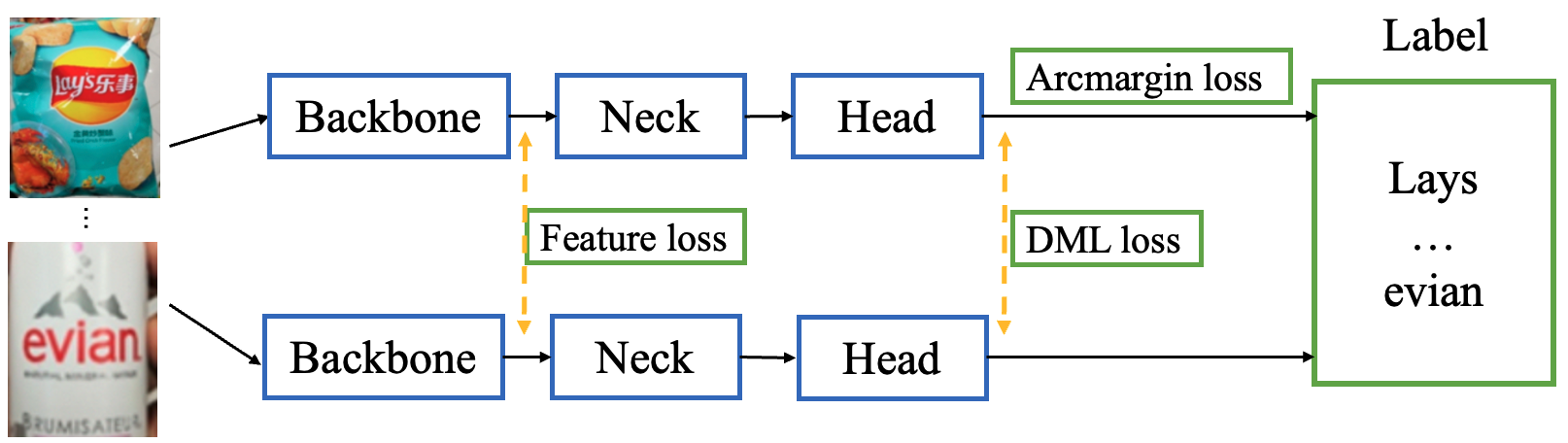}
}
\caption{U-DML framework}
\label{udml_framework}
\end{figure*}

To avoid too time-consuming teacher model training process, in this paper, based on DML, we proposed U-DML, in which feature maps are also supervised during the distillation process. Figure \ref{udml_framework} shows the framework of U-DML.

There are two networks for the distillation process: the student network and the teacher network. They have exactly the same network structures with different initialized weights. The goal is that for the same input image, the two networks can get the same output, not only for the prediction result but also for the feature map.

The total loss function consists of three parts: (1) Arcmargin loss. Since the two networks' neck and head are trained from scratch, Arcmargin loss can be used for the networks' convergence; (2) DML loss. The final output distribution of the two networks is expected to be the same, so DML loss is needed to ensure the consistency of the distribution between the two networks; (3) Feature loss. The architectures of the two networks are the same, so their feature maps are expected to be the same, feature loss can be used to constrain the two networks' intermediate feature map distance.

\begin{small}
\begin{equation}
Loss_{arc} = Arcmargin(S_{hout}, gt) + Arcmargin(T_{hout}, gt)
\end{equation}
\label{loss_arc}
\end{small}

in which $S_{hout}$ denotes head output of the student network and $T_{hout}$ denotes that of the teacher network. $gt$ donates the groudtruth label of the input image.

\textit{DML loss.}
In DML, parameters of each sub-network are updated separately. Here, to simplify the training process, we calculate the KL divergence loss between the two sub-networks and update all the parameter simultaneously. The DML loss is as follows.

\begin{small}
\begin{equation}
Loss_{dml} = \frac{KL(S_{pout} || T_{pout}) + KL(T_{pout} || S_{pout})}{2}
\end{equation}
\label{loss_dml}
\end{small}

in which $KL(p || q)$ denotes KL divergence of the $p$ and $q$. $S_{pout}$ and $T_{pout}$ can be calculated as follows.

\begin{small}
\begin{equation}
\begin{aligned}
S_{pout} = Softmax(S_{hout}) \\
T_{pout} = Softmax(T_{hout})
\end{aligned}
\end{equation}
\label{pout_calc}
\end{small}

\textit{Feature loss.}
During the training process, we hope that the backbone output of the student network is the same as that of the teacher network. Therefore, similar to Overhaul, feature loss is used for the distillation process. The loss can be calculated as follows.

\begin{small}
\begin{equation}
Loss_{feat} = L2(S_{bout}, T_{bout})
\end{equation}
\label{loss_feat}
\end{small}

in which $S_{bout}$ means backbone output of the student network and $T_{bout}$ means that of teacher network. Mean square error Loss is utilized here. It is noted that for the feature loss,  feature map transformation is not needed because the two feature maps used to calculate the loss are exactly the same.

Finally, the total loss for the U-DML training process is shown as follows.

\begin{small}
\begin{equation}
Loss_{total} = Loss_{arc} + Loss_{dml} + Loss_{feat}
\end{equation}
\label{loss_udml_total}
\end{small}

\subsubsection{DeepHash}

In practical applications, the massive retrieval images and videos feature database will not only consume huge storage space, but also result in long retrieval time, which is unacceptable in some application scenarios. In addition, the storage space on the mobile device is very limited, which puts forward further challenges for massive feature storage. With this in mind, we integrate DeepHash \cite{2020A} functionality in addition to Metric Learning capabilities in PP-Shitu which can help to decrease storage and speed up retrieval

DeepHash studies how to obtain representative binary features using deep neural networks, which uses bit to store binary features, and adopt Hamming distance to measure the distance between two feature vectors. Specifically, we have open sourced three of the most common DeepHash algorithms: DLBHC  \cite{2015Deep}, LCDSH \cite{2017HashNet}, DSHSD \cite{2019Deep}.

In PP-ShiTu, we adopt the optimized DSHSD \cite{2019Deep} algorithm to get practical binary feature for identifying goods. Compared with the real-valued model, the retrieval precision may slightly drop, but the storage space of gallery features can be reduced 32 times, and the retrieval speed can be increased by more than 5 times when the size of the search library larger than 100 thousand. 

\subsection{Vector Search}

In order to run on multiple operating system, including Linux, Windows and MacOS, we use Faiss \footnote{https://github.com/facebookresearch/faiss} as the vector search module, which is a library for efficient similarity search and contains algorithms that support search in sets of vectors of any size, up to ones that possibly do not fit in RAM. In PP-ShiTu, we choose three algorithms, HNSW32 \cite{malkov2018efficient}, IVF \cite{6248038}, FLAT to meet the demand of different scenarios. HNSW32 is a hugely popular approximate nearest neighbor algorithm that produces state-of-the-art performance with super fast search speeds and fantastic recall. Because of based on graph, HNSW32 only supports add element after the graph created. IVF is an index data structure storing a mapping from content. Although it is not as effective as HNSW32, IVF supports deleting elements after building index. The FLAT is a violent retrieval algorithm with highest accuracy, but it is the slowest compared with the first two methods.

It is worth noting that a feature gallery needs to be built before vector search. The feature gallery is composed of features extracted from labeled images. In PP-ShiTu pipeline, the label of query is obtained from feature gallery after similarity search.

\section{Experiments}

\subsection{Experimental Setup}

\subsubsection{Datasets}
We used a lot of public datasets for different scenarios to train our model, the specific introduction is as follows.

\textbf{Mainbody Detection}
For mainbody detection training process, we use 5 public datasets: Objects365 \cite{shao2019objects365}, COCO2017 \cite{lin2014microsoft}, iCartoonFace \cite{zheng2020cartoon}, LogoDet-3k \cite{wang2020logodet3k} and RPC \cite{wei2019rpc}.  More detailed information can be seen in Table \ref{mainbody_dataset}.

\begin{table*}[!h]
\begin{center}
\begin{tabular}{c|c|c|c|c}
\toprule[1pt]
Dataset  & \makecell[c]{Number of \\ training images } & \makecell[c]{Images used \\ for training} & \makecell[c]{Images used \\ for validation}  & Scenarios  \\
\midrule[1pt]
Objects365 &  1700K & 173k & 6K & General Scenarios \\
COCO2017 & 118K & 118k & 5K & General Scenarios \\
iCartoonFace &  48k & 48k & 2K & Cartoon Face \\
LogoDet-3K & 155k & 155k & 2K & Logo  \\
RPC & 54k & 54k & 3k & Product \\
\bottomrule[1pt]
\end{tabular}
\end{center}
\caption{Statement of mainbody detection training dataset}
\label{mainbody_dataset}
\end{table*}

\textbf{General Recognition}
For general recognition, we use 7 public datasets including Aliproduct \cite{cheng2020weakly}, GLDv2 \cite{weyand2020google}, VeRI-Wild \cite{lou2019large}, LogoDet-3K \cite{wang2020logodet3k}, iCartoonFace \cite{zheng2020cartoon}, SOP \cite{songCVPR16}, Inshop \cite{liu2016deepfashion}. These data sets are also used for general training, model distillation and deephash. The detail of each dataset is described in the Table \ref{rec_dataset}.

\begin{table}[!h]
\begin{center}
\begin{tabular}{c|c|c|c}
\toprule[1pt]
Dataset  &  \makecell[c]{Number of\\images} &   \makecell[c]{Number of\\classes} & Scenarios \\
\midrule[1pt]
Aliproduct &  2498771 & 50030 & Product\\
GLDv2 & 1580470 & 81313 & Landmark\\
VeRI-Wild & 277797& 30671 & Vehicle\\
LogoDet-3K & 155427 & 3000 & Logo  \\
iCartoonFace & 389678 & 5013 & Cartoon\\
SOP & 59551 & 11318 & Product \\
Inshop & 25882 & 3997 & Product\\
\hline
\textbf{Total} & \textbf{5M} & \textbf{185K} & ——\\
\bottomrule[1pt]
\end{tabular}
\end{center}
\caption{Statement of recognition training dataset}
\label{rec_dataset}
\end{table}

\begin{figure*}[h]
\centering
\includegraphics[width=12cm]{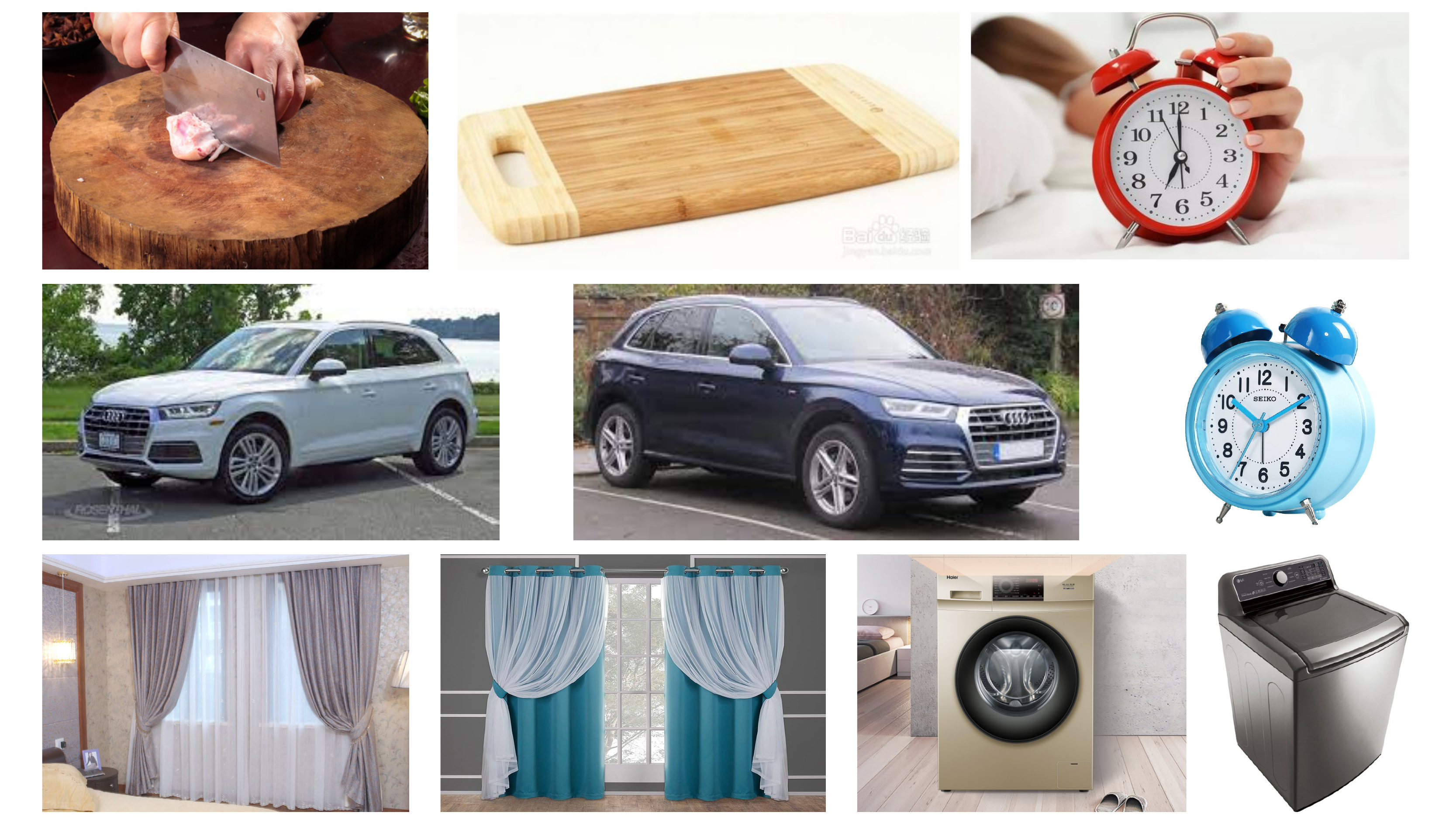}
\caption{Examples of images in test set.}
\label{examples}
\end{figure*}

\textbf{Test set}
To evaluate the speed of PP-ShiTu, we release a test set consisting of a query image set and a feature library. We collect 200 images from Internet as the query set. The query dataset can be found in the GitHub repository PaddleClas. For gallery, we used a combination of Products-10k\cite{bai2020products} and some common products downloaded from the Internet, and build feature library with image gallery. Each image in the query set contains at least one object from gallery. Figure \ref{examples} shows the examples of images in test set.

\subsubsection{Implementation Details}

The following briefly describes our implementation details, which are critical to the result of the final system.

\textbf{Mainbody Detection}
For mainbody detection, we use PP-PicoDet architecture as our mainbody detection architecture. To improve the backbone's feature extraction capabilities, we use PP-LCNet-2.5x as the detector backbone, and the pretrained weights of PP-LCNet-2.5x is trained with SSLD distillation method \cite{cui2021selfsupervision} using PaddleClas.

To validate the effectiveness of PP-LCNet as PP-PicoDet's bacbone, we carry out the experiment on COCO2017 dataset firstly. The batch size and learning rate are reduced to half of the standard configuration provided in PaddleDetection because of the GPU memory limit.

For the mainbody detection, the total epoch number for mainbody detection is 100 and the other hyper-parameters are same as that on COCO2017 dataset before.

\textbf{General Recognition}
For general recognition, we use all 7 datasets mentioned above to train our PP-LCNet-2.5x for 100 epochs. We follow standard practice and perform data augmentation with random-size cropping to 224$\times$224 pixels and random horizontal flipping. Optimization is performed using SGD with momentum 0.9 and a mini-batch size of 2048. Weight decay is empirically set to 1e-5 and initial learning rate is 0.1. In Arcmargin Loss, we set the scale to 30 and the margin to 0.2. In order to further improve the indicators in small datasets, we have increased the sampling rate of LogoDet, Inshop, and VeRI-Wild by four times. The hyper-parameters of U-DML distillation training are the same with those of the general recognition except the loss function. In the evaluation phase, we use each single dataset for evaluation, and the metric we use is recall@1. In DSHSD Loss, the weight of contrastrive loss is set to 0.05, and  the margin is set to 1024 (twice of the feature embedding size).

\subsection{Mainbody Detection Result}

Table \ref{picodet_coco_result} shows the results of PP-LCNet as PP-PicoDet's backbone. It shows the the model has a huge advantage over PP-YOLOv2-ResNet50vd-DCNv2 model on model size. And in the case of similar model size, its mAP result is alsw much higher than PP-YOLO tiny model.

\begin{table*}[!h]
\begin{center}
\begin{tabular}{c|c|c|c|c}
\toprule[1pt]
Architecture & Backbone  &  Input size &   mAP@0.5:0.95  & Model size(MB) \\
\midrule[1pt]
PP-YOLOv2 & ResNet50\_vd-DCNv2 & 640x640 & 49.1 & 250 \\
PP-YOLO tiny & MobileNetV3-large-1.0x & 320x320 & 23.2 & 28.0 \\
PP-PicoDet &  PP-LCNet-2.5x & 640x640 & 41.0 &  29.2 \\
\midrule[1pt]
\end{tabular}
\end{center}
\caption{COCO2017-val mAP for different models}
\label{picodet_coco_result}
\end{table*}

Table \ref{detection_exp_result} shows the mAP results on 5 public data sets. PP-YOLO is the mainbody detection model we provided in the last version, which is suggested for server-side usage. PP-PicoDet is 14 times faster than PP-YOLO on CPU latency while just has a slight mAP reduction (-2.4\%), To further reduce the model size, we quantize PP-PicoDet model using PACT and the final model size is reduced to 6.9M, which is suitable for mobile deployment.

\begin{table*}[h!]
\begin{center}
\begin{tabular}{c|c|c|c}
\toprule[1pt]
-  & PP-YOLO R50-vd-DCNv2 &  PP-PicoDet & PP-PicoDet Int8 \\
\midrule[1pt]
Objects365 &  0.427 & 0.382 & 0.387\\
COCO2017 & 0.480 & 0.444 & 0.450\\
iCartoonFace & 0.399 & 0.371 &  0.389\\
LogoDet-3K & 0.507 & 0.575 & 0.601\\
RPC & 0.312 & 0.570 & 0.548\\
\hline
Average & 0.425 & 0.468 & 0.475 \\
\hline
Model size (MB) & 250 & 29.2 & --\\
\hline
Latency on CPU (ms) & 466.0 & 29.8 & --\\
\bottomrule[1pt]

\end{tabular}
\end{center}
\caption{Ablation experiments of mainbody detection models in PP-ShiTu. Latency tested on Intel$^\circledR$ Xeon$^\circledR$ Gold 6148 Processor with batch size of 1 and  MKLDNN enabled, the number of thread is 10.}
\label{detection_exp_result}
\end{table*}

\subsection{General Recognition Result}
Table \ref{rec_model_recall} shows the recall@1 of ResNet50\_vd and PP-LCNet-2.5x on different dataset. Among them, ResNet50\_vd is trained on their respective datasets. PP-LCNet-2.5x is trained on the entire merged dataset.
As can be seen from the table, our training strategy allows our lightweight model to approach or even exceed the accuracy of ResNet50\_vd. Moreover, because our recognition model is trained on a large number of different types of data, it can be applied to different scenarios.

\begin{table*}[h!]
\begin{center}
\begin{tabular}{c|c|c|c|c|c|c|c|c}
\toprule[1pt]
  & \multicolumn{7}{c|}{Recall@1} \\
\cline{3-8}
 Model & Train dataset &Aliproduct & VeRI-Wild & LogoDet-3k & iCartoonFace & SOP & Inshop & \makecell[c]{Latency\\(ms)}\\

\midrule[1pt]
ResNet50\_vd    & Aliproduct  &  \textbf{0.836}  & 0.234   & 	
0.792 & 0.465 & 0.63262 & 0.405& \\
ResNet50*     & VeRI-Wild &  	0.063  & \textbf{0.938}   & 0.575 & 0.154 & 0.233 & 0.071 &15.2  \\
ResNet50*    & LogoDet-3k  &  0.097  &  0.118  & \textbf{0.898} & 0.191 & 0.236 & 0.061&  \\
ResNet50\_vd    & iCartoonFace  &  0.099 & 0.068   & 0.694 & \textbf{0.832} & 0.543 & 0.253 &  \\
ResNet50\_vd    & Inshop  &  0.187 & 0.075   & 0.712 & 0.308 & 	
0.621 & 0.887&  \\
\hline
\textbf{PP-LCNet-2.5x}   & All  &  \textbf{0.839}  & 0.888   & 0.861 & \textbf{0.841} & \textbf{0.793} & \textbf{0.892}& 5.0   \\ 
\hline
\textbf{ \makecell[c]{PP-LCNet-2.5x \\Int8}}   & All  &  0.819  & 0.847   & 0.865& 0.821 & 0.755 & 0.870 & -   \\ 
\bottomrule[1pt]
\end{tabular}
\end{center}
\caption{Comparison of ResNet50 and PP-LCNet-2.5x in PP-ShiTu on each dataset. * indicates that the stride of the last block in the backbone is 1, which means larger FLOPs. Latency tested on Intel$^\circledR$ Xeon$^\circledR$ Gold 6148 Processor with batch size of 1 and  MKLDNN enabled, the number of thread is 10.}
\label{rec_model_recall}
\end{table*}

\subsection{Ablation study}
\subsubsection{Ablation study for PP-LCNet}
To test the generalization ability of the model, we use challenging datasets like ImageNet-1k throughout the process of designing the model. Table \ref{PP-LCNet-Ablation} shows the accuracy-speed comparison between PP-LCNet and other competive lightweight models. It can be seen that our PP-LCNet has a significant advantage both on speed and accuracy, even when compared to a very competitive network like MobileNetV3.

\begin{table*}[h!]
\centering
\begin{center}
\begin{tabular}{c|c|c|c|c|c}
\toprule[1pt]
Model & Params(M) & FLOPs(M) & Top-1 Acc.(\%) & Top-5 Acc.(\%) & Latency(ms) \\
\midrule[1pt]
MobileNetV2-0.25x     & 1.5     & 34     & 53.21	 & 76.52 & 2.47 \\
MobileNetV3-small-0.35x    & 1.7     & 15     & 53.03 & 76.37  & 3.02 \\
ShuffleNetV2-0.33x & 0.6     & 24     & 53.73  & 77.05  & 4.30 \\

\textbf{PP-LCNet-0.25x}    & \textbf{1.5}    & \textbf{18}    & \textbf{51.86}  & \textbf{75.65} & \textbf{1.74} \\
\hline
MobileNetV2-0.5x      & 2.0     & 99     & 65.03 & 85.72 & 2.85 \\
MobileNetV3-large-0.35x & 2.1 & 41  & 64.32  & 85.46  & 3.68 \\
ShuffleNetV2-0.5x     & 1.4     & 43 & 60.32    & 82.26  & 4.65 \\ 
\textbf{PP-LCNet-0.5x} & \textbf{1.9}  & \textbf{47}   & \textbf{63.14}      & \textbf{84.66}  &\textbf{2.05}\\ 

\hline
MobileNetV1-1x  & 4.3   & 578     & 	70.99 & 89.68 & 3.38 \\
MobileNetV2-1x  &  3.5  & 327     & 72.15 & 90.65 & 4.26 \\
MobileNetV3-small-1.25x & 3.6 & 100  & 70.67  & 89.51  & 3.95 \\
ShuffleNetV2-1.5x   & 3.5   & 301 & 71.63    & 90.15  & - \\ 
\textbf{PP-LCNet-1x}   & \textbf{3.0}  & \textbf{161}  & \textbf{71.32}    & \textbf{90.03}    &\textbf{2.46}\\ 

\bottomrule[1pt]
\end{tabular}
\end{center}

\caption{PP-LCNet ablation study. Comparison of state-of-the-art light networks over classification accuracy. Latency tested on Intel$^\circledR$ Xeon$^\circledR$ Gold 6148 Processor with batch size of 1 and  MKLDNN enabled, the number of thread is 10.}
\label{PP-LCNet-Ablation}
\end{table*}

\subsubsection{Ablation study for U-DML}

Based on the general PP-LCNet-2.5x recognition model, we carry out some experiments. The training strategy is almost the same with that for base PP-LCNet-2.5x recognition model training process, except that we adopt distillation loss. Table \ref{distillation_rec_model_recall} shows some results for different methods.

Model recall@1 on 6 data sets is improved by 0.5\% on average using DML, and after using U-DML, the model recall@1 is improved by 0.7\%. Since the model is generated using knowledge distillation training process, it will not take any extra time cost during inference.

It is noted that we did not fine-tune any hyper-parameters for the model training process such as loss ratio. What's more, compared to the standard training process (the final train loss is 2.64), the distillation model tends to generalize better and takes longer time to converge (the final train loss of the student is 2.75), which is consistent with the conclusion in \cite{beyer2021knowledge}. We believe that with more epochs, the distillation model can perform better.

\subsubsection{Ablation study for DeepHash}

Based on DSHSD algorithm, we trained an binary model with same backbone, embedding size, train strategy and dataset as real-valued. Also, We carry out some experiments compared with real-valued model on precision, index size and retrieval time. The Ali-product val dataset which consists of 145235 images is adopted as the gallery dataset. Table \ref{deep_hash_ablation} shows the results of these two models. We can see that real-valued model has higher retrieval precision, while binary model consumes few storage and retrieval time. 

\begin{table*}[h!]
\begin{center}
\begin{tabular}{c|c|c|c|c|c}
\toprule[1pt]
& \multicolumn{3}{c|}{Recall@1} & \\
\cline{2-4}
Model Type & Ali-Product & iCartoonFace& LogoDet-3K & \makecell[c] Index Size(M) & Retrieval Time(ms)\\
\midrule[1pt]
Real-Value &	\textbf{0.839} & \textbf{0.832} & \textbf{0.868} &285 &117\\
Binary & 0.78 &	0.775 & 0.833 & \textbf{8.9} & \textbf{21} \\
\bottomrule[1pt]
\end{tabular}
\end{center}
\caption{Ablation experiments of binary model and  real-valued model. The retrieval time is tested with the method of Flat}
\label{deep_hash_ablation}
\end{table*}

\subsection{Inference time latency on test set}
We compare inference speed of PP-ShiTu and the server model, which was released on PaddleClas earlier and use ResNet 50 backbone for detection and feature extraction, on the test set. F1-scores of two models are both 0.3306. Table \ref{system_latency} shows the latency of two models tested on CPU and GPU server.

\begin{table*}[h!]
\begin{center}
\begin{tabular}{c|c|c|c|c|c|c|c}
\toprule[1pt]
  & \multicolumn{7}{c}{Recall@1} \\
\cline{2-8}
Model &Aliproduct & VeRI-Wild & LogoDet-3k & iCartoonFace & SOP & Inshop & Average \\
\midrule[1pt]
PP-LCNet-2.5x    &  0.839  & 0.888   & 0.861 & 0.841 & 0.793 & 0.892 & 0.852   \\

PP-LCNet-2.5x + DML    &  0.841  & 0.892   & 0.863 & 0.848 & 0.803 & 0.894 & 0.857   \\
\textbf{PP-LCNet-2.5x + U-DML }    &  \textbf{0.841}  & \textbf{0.901}   & \textbf{0.863} & \textbf{0.849} & \textbf{0.803} & \textbf{0.896} & \textbf{0.859}   \\
\bottomrule[1pt]
\end{tabular}
\end{center}
\caption{Ablation experiments of knowledge distillation of recognition model}
\label{distillation_rec_model_recall}
\end{table*}

\begin{table*}[h!]
\begin{center}
\begin{tabular}{c|c|c|c|c|c}
\toprule[1pt]
\multirow{2}{*}{Processor Info. } & \multirow{2}{*}{Model} 
& \multicolumn{4}{c}{Latency (ms)} \\
\cline{3-6}
&& Detection & Feature Extraction & Vector Search & Overall  \\
\midrule[1pt]
\multirow{2}{*}{Intel\textregistered Xeon\textregistered Gold 6148} & Server Model & 923&79&44&1051\\
& \textbf{PP-ShiTu} & \textbf{94} & \textbf{45} & 51 & \textbf{194}\\
\cline{1-6}
\multirow{2}{*}{Intel\textregistered Core\textsuperscript{TM} i5-6300U } & Server Model & 3305&475&3.3&3790\\
& \textbf{PP-ShiTu } & \textbf{282} & \textbf{150} & 4.4 & \textbf{442}\\
\bottomrule[1pt]
\end{tabular}
\end{center}
\caption{Inference time of PP-ShiTu and server model on test set. Latency tested on models with batch size of 1 and  MKLDNN enabled, the number of thread is 10.}
\label{system_latency}
\end{table*}

\section{Conclusions}
In this paper, we propose a practical lightweight image recognition system, PP-ShiTu. We introduce a general pipeline and practical strategies including PP-PicoDet, U-DML, Arcmargin loss and DeepHash. Experiments show that PP-ShiTu works very well on different tasks and datasets. The corresponding ablation experiments are also provided.

\bibstyle{aaai21}
\bibliography{eg}
\end{document}